\documentclass[10pt,journal,compsoc]{IEEEtran}
\ifCLASSOPTIONcompsoc
  \usepackage[nocompress]{cite}
\else
  \usepackage{cite}
\fi
\ifCLASSINFOpdf
\else
\fi
\usepackage[pagebackref=false,breaklinks=true,letterpaper=true,colorlinks,bookmarks=false]{hyperref}

\usepackage{times}
\usepackage{epsfig}
\usepackage{graphicx}
\usepackage{amsmath}
\usepackage{amssymb}
\usepackage{booktabs}
\usepackage{rotating}
\usepackage{multirow}
\usepackage{lipsum}

\usepackage{amsfonts}
\usepackage{microtype}
\usepackage{color}
\usepackage{hyperref}
\usepackage{mdwlist}
\usepackage{pbox}
\usepackage{booktabs}

\usepackage{eso-pic}
\usepackage{xspace}

\def\figurePath{figures/}
\def\myfigure#1#2{\begin{figure}[ht]\centering\includegraphics*[width = \linewidth]{\figurePath#1}\caption{#2}\label{fig:#1}\end{figure}}
\def\mycfigure#1#2{\begin{figure*}[tb]\centering\includegraphics*[clip, width = \linewidth]{\figurePath#1}\caption{#2}\label{fig:#1}\end{figure*}}

\def\mysection#1#2{\section{#1}\label{sec:#2}}

\def\mysubsection#1#2{\subsection{#1}\label{sec:#2}}
\def\mysubsubsection#1#2{\subsubsection{#1}\label{sec:#2}}

\newcommand{\refSec}[1]{Sec.~\ref{sec:#1}}
\newcommand{\refFig}[1]{Fig.~\ref{fig:#1}}
\newcommand{\refEq}[1]{Eq.~\ref{eq:#1}}
\newcommand{\refTbl}[1]{Tbl.~\ref{tbl:#1}}

\makeatletter
\renewcommand{\paragraph}{%
  \@startsection{paragraph}{4}%
  {\z@}{2.25ex \@plus 1ex \@minus .2ex}{-1em}%
  {\normalfont\normalsize\bfseries}%
}

\DeclareRobustCommand\onedot{\futurelet\@let@token\@onedot}
\def\@onedot{\ifx\@let@token.\else.\null\fi\xspace}

\def\eg{\emph{e.g}\onedot} 
\def\ie{\emph{i.e}\onedot} 
 
\def\etc{\emph{etc}\onedot} 
\def\wrt{w.r.t\onedot} 
\def\etal{\emph{et al}\onedot}
\makeatother

\begin{document}
\title{%
Novel Views of Objects from a Single Image
}
%
%
%
%

\author{Konstantinos~Rematas,
        Chuong~Nguyen,
        Tobias~Ritschel,
        Mario~Fritz,
        and~Tinne~Tuytelaars
\IEEEcompsocitemizethanks{
\IEEEcompsocthanksitem Project \url{http://homes.cs.washington.edu/\~krematas/ViewSynthesis/}
}

}

%
%

\markboth{}%
{}
%



\IEEEtitleabstractindextext{%
\begin{abstract}

Taking an image of an object is at its core a lossy process. The rich information about the three-dimensional structure of the world is flattened to an image plane and decisions such as viewpoint and camera parameters are final and not easily revertible. As a consequence, possibilities of changing viewpoint are limited.
Given a single image depicting an object, novel-view synthesis is the task of generating new images that render the object from a different viewpoint than the one given. The main difficulty is to synthesize the parts that are disoccluded; disocclusion occurs when parts of an object are hidden by the object itself under a specific viewpoint.
In this work, we show how to improve novel-view synthesis by making use of the correlations observed in 3D models and applying them to new image instances. We propose a technique to use the structural information extracted from a 3D model that matches the image object in terms of viewpoint and shape. For the latter part, we propose an efficient 2D-to-3D alignment method that associates precisely the image appearance with the 3D model geometry with minimal user interaction. Our technique is able to simulate plausible viewpoint changes for a variety of object classes within seconds. Additionally, we show that our synthesized images can be used as additional training data that improves the performance of standard object detectors.

\end{abstract}

}

\maketitle

\IEEEdisplaynontitleabstractindextext

%
\IEEEpeerreviewmaketitle

\IEEEraisesectionheading{\section{Introduction}\label{sec:introduction}}

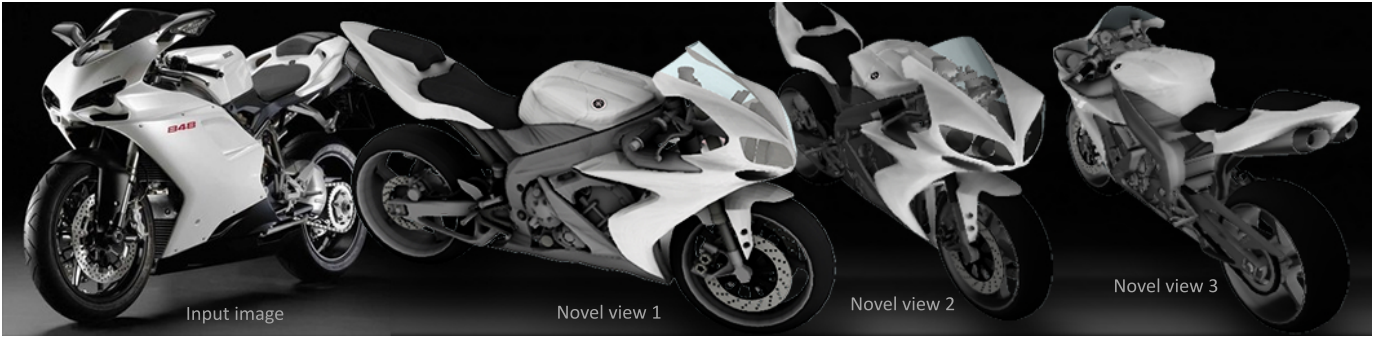
\mycfigure{Teaser}{Given an input image \emph{(left)}, we synthesize novel views of the depicted object at interactive frames rates with view-dependent apearance \emph{(right)}.}

\IEEEPARstart{G}{iven} a single image of an object, humans can clearly ``imagine" what this object would look like from different viewpoints. Even when the particular object was never seen before, its visible parts are enough to facilitate the imaginary synthesis that happens in our mind. For example when we see the side view of a car, we can hallucinate the front part, including details such as lights, hood, wind shield etc.
What enables us to do so? The answer lies in our knowledge of the structural information for a particular visual class. The object instances of that class have specific 3D shapes, symmetries, materials, colors, \etc. Several of these attributes are shared among the class instances and thus they can be used as cues for the synthesis process. Probably the most important cues for synthesis are those related with the 3D structure of the objects. The reason is that this task shares the same principles with image formation: an image is not just pixels in a grid but it is the 2D interpretation of a 3D world.

The synthesized view needs to be a plausible simulation of how the object would appear in the new view. There are two parameters in such a setting: a) the underlying geometry and b) the associated appearance. The novel view needs to have the geometric characteristics of the object's class and the appearance of the depicted object instance. If only one of the two is inconsistent then the result is perceived as fake or unrealistic. For example, simple replacement of a car with a rendered car 3D model keeps the geometric structure of the class, but the appearance will come from the 3D model and may not blend in well. On the other hand, if we discard the unseen geometry then in a new view the object will have empty areas, similar to a facade in a Potemkin village.

The ability to synthesize novel views opens new possibilities in several image manipulation tasks. A direct application is the 3D manipulation of objects in images. In addition, the novel views can be used as additional training data for conventional object detectors/classifiers. The emergence of 3D or light field or stereo displays provide another domain that benefits from the generation of additional views of objects. Uncovering the relationship between appearance and geometry, a necessary step for the view synthesis, can help in tasks such as shape manipulation, appearance transfer, \etc. Having information about the 3D structure of an object is essential also for image understanding. After all, an object is not just a two-dimensional box but rather a real world entity with certain geometry, materials and interaction with the environment.

The work presented in this paper focuses on the synthesis of new views of objects from single images and the necessary prerequisites to achieve this. We approach the synthesis problem by considering the 3D models of a class as proxies to transfer the appearance to the novel views. We start by assuming manual alignment between the 2D appearance and the 3D shape of the object, and we synthesize directly the pixels in the novel view as a combination of pixels in the original view. Next, we extend our approach to a user-centered setting, where interaction and speed are essential, and viewpoint-dependency of appearance is taken into account. We present a method for aligning the geometry and the appearance with minimal user interaction. Then, we revisit the view synthesis and we extend it to include viewpoint dependency for the reconstruction in an efficient way.
Our key contributions can be summarized as:
\begin{itemize}
\item A method to synthesize realistic novel 3D views of an object from a single 2D image
\item A procedure to align a 3D shapes to a single image
\item Amplification of training data in detection and classification tasks by novel-view synthesis.
\end{itemize}

This article is organised as follows: the previous work is described in~\refSec{PreviousWork}.
In~\refSec{NovelViewSynthesis} we introduce our core method for novel view synthesis.~\refSec{Alignment} focuses on the matching between the geometry and the appearance of the object. In~\refSec{ViewSynthesis2} we revisit the novel view synthesis from a viewer-centered perspective. Experimental results are reported in Section~\refSec{Experiments} and further discussed in Section~\refSec{Discussion}.
Section~\refSec{Conclusion} concludes the paper.

\mysection{Previous work}{PreviousWork}
\paragraph*{View Synthesis}
Generating novel views from 2D images is an image-based rendering problem, with several applications in computer vision \cite{Beier1992} and computer graphics  \cite{Chen1993,Horry1997}.
Image-based rendering techniques \cite{Gortler1996} can convincingly synthesize novel viewpoints with complex material and illumination effects from a set of images.
It has been shown that quality can greatly benefit if a notion of 3D geometry is provided to the system \cite{Debevec1996}.
Kholgade~\etal\cite{Kholgade2014} showed that given a manual 3D shape alignment convincing and detail-preserving novel views can be generated even from a single image.
View interpolation including complex shading is possible by using surface light fields \cite{Wood2000} or lumigraphs \cite{Gortler1996}.
When deforming and blending images, detecting holes and filling them is the biggest challenge \cite{Debevec1998}, usually solved either by stretching foreground over background or by inpainting \cite{Criminisi2003,BarnesSIG2009}, but no approach we are aware of uses the structure of a 3D template to guide this inpainting.
Appearance that depends on a shape template was proposed for the diffuse case and manual intervention to capture the appearance of art in the ``Lit Sphere'' metaphor \cite{Sloan2001}.
Our approach can be seen as the automatic and continuous generalization from one to many, context-dependent Lit Spheres.
Image Analogies \cite{Hertzmann2001} learns the transformation from a pair of images, to apply to a novel image.
Similarly, we learn the relation between two rendered 3D views of an image to generate appearance in new views from a single image for general appearance and account for the relation of 3D geometry and materials \cite{Jain2012}.
Simpler, joint bilateral upsampling \cite{Kopf2007} has been used to reconstruct a high-resolution image from a low-resolution image using a high-resolution guidance signal.
Our approach uses multiple synthesized high-resolution images to guide view synthesis, including upsampling.
For 3D reconstruction of occluded areas in scanned objects, approaches based on templates are routinely used \cite{Pauly2005} to fill holes in surfaces, but these operate on 3D surfaces and do not account for appearance such as we do.
Another approach to reconstruct occluded regions is \cite{vandenHengel07}, but the method requires visibility of the region or its symmetric part in the video sequence.
Commonly, acquiring view-dependent appearance in terms of a BRDF requires to also capture light probe, even when the geometry is known \cite{Wu2015}.
We circumvent this requirement by not factoring out reflectance and lighting, while still allowing for plausible view-dependent appearance (such a decomposition can be achieved using \cite{RematasArxiv2016}). More recently, the work of \cite{RematasCVPR16} estimates the appearance of an object using Reflectance Maps~\cite{Horn1979} but requires single material objects.

\paragraph*{2D-to-3D alignment}
If the object belongs to a parametric family of shapes, alignment can be automated \cite{Blanz1999}.
For cars and bicycles, Zia~\etal\cite{ZiaPAMI2013} have shown accurate alignments with a wireframe model based on a deformable 3D shape model and robust part detectors, starting from CAD models with manual part annotations.
The situation is more difficult if no parametric model is available. The recent work of \cite{Huang2015} performs 3D reconstruction of objects in image collections, but their method is not interactive as it requires the images to be processed jointly.
When a shape deformation model is available, the 3D shape can be deformed to fitting the 2D silhouette also leading to a 2D-to-3D alignment \cite{Xu2011}.
We combine silhouette and luminance pattern-based approaches in one discriminative learning framework.

A recent suggestion is to use 3D shapes and extract discriminative patches \cite{Doersch2012} to find the alignment, \eg to paintings \cite{Aubry2013} or single objects \cite{Aubry2014,shmlg_imageDepth_sig14}.
Such an approach has been shown able to reliably find instances from a class for which a large set of 3D shapes is available. They obtain good precision, in real images with clutter, but at high computational cost \cite{Aubry2014}. Lim~\etal\cite{Lim13} propose a RANSAC-based sampling approach in order to speed up the search - but this only works for exact 3D shapes and cannot cope with within-class variability.
Using templates, it can require up to 2 minutes to align a single chair in a cluster of 80 cores, which is far from being usable in an interactive application \cite{Aubry2014}.
The results in this article are computed an order of magnitude faster while achieving higher quality.
Our approach is also related to the recent work of \cite{Choy2015} in terms of extending the fine view alignment step to 3D shape parts to allow for shape deformations.

\paragraph*{Training from synthetic data}
There is an increasing interest to train computer vision models from synthesized data in order to achieve more scalable learning schemes.
Work on fully synthetic data has shown mostly successful in depth data \cite{shotton11cvpr,lai09rss,wohlkinger10cogsys}, shape-based representations \cite{stark10bmvc,liebelt10cvpr}, textures \cite{TarghiICPR2008,li12eccv} and scenes \cite{kaneva11iccv}. In contrast, we take an image-based approach in order to leverage real-world image statistics as well as the intra-class variation available to us in image data.
Previous image-based work synthesizes training images by recombining poses and appearance of objects in order to create new instances \cite{enzweiler08cvpr,leonid11cvpr, pishchulin11bmvc,pepik12cvpr,2002ZobelObjec}.
More recently, Su~\etal\cite{Su_2015_ICCV} and Peng~\etal\cite{Peng_2015_ICCV} rendered 3D models to generate additional data for object detection and pose estimation respectively using Convolutional Neural Networks. In contrast, our work focuses on synthesis across viewpoints with the appearance given from an image and deals with disocclusions that are not addressed in previous work.

\mysection{3D Model-guided Novel-view synthesis}{NovelViewSynthesis}
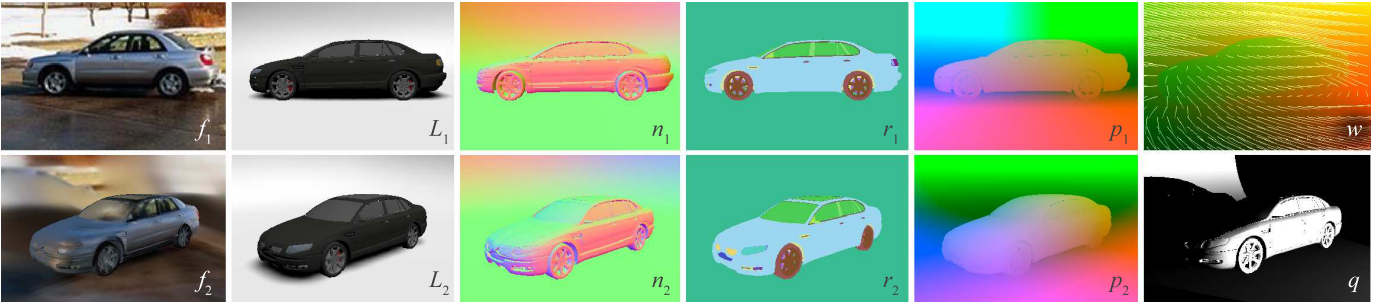
\mycfigure{Overview}{We compute a novel view image $f_2$ from the input image $f_1$ using 3D information of an aligned 3D template (Left to right: radiance, normals, reflectance, positions) as guidance, even if its renderings $L_1$ and $L_2$ have appearance largely different from $f_1$. $q$ is the sampling quality, indicating the visibility in the original view, and $w$ is the flow between the views, which is optional and used to speed up the sampling.}

\textit{We pose novel view synthesis as a problem of reconstructing appearance as a function of scene attributes from observed samples.}
Input to our algorithm is an image and an aligned 3D model and output is an image from a novel view.
Firstly, the original and the novel view of the 3D model can be rendered easily.
This allows to put pixel locations from the original to the novel view into correspondence (flow) and to detect disocclusions, \ie, pixels in the novel view that were not present in the original view.
Simply copying pixels along their flow will result in the classic projective texturing used in image-based rendering \cite{Debevec1998}.
The challenge is to fill the disocclusions by inpainting.
The most simple solution would be to replace disoccluded pixels with pixels from the rendered 3D model.
This, however, cannot produce a consistent appearance as precise materials, lighting and geometry are unknown.

The key observation is that the 3D model provides several 3D properties, such as position, normal, reflectance, for all disoccluded pixels (\refFig{Overview}).
We can use this information to guide the inpainting of appearance for such pixels.
To this end, we copy appearance from the original view, that is similar in terms of position, normal and reflectance.
Briefly, each pixel in the novel view can be modeled as a weighted sum of the pixels in the original view. These weights are essentially a compatibility score between the position, normal and reflectance values of pixels in the two views.

\mysubsection{Sampling appearance}{SamplingAppearance}
Input to our system is appearance in the form of a 2D RGB image $f_1: \mathbb R^2\rightarrow\mathbb R^3$ and a 3D model $\mathcal S\subseteq\mathbb R^3$ with Phong reflectance $\mathcal F:\mathcal S\rightarrow\mathbb R^9$ defined on it (diffuse color, specular color and glossiness).
We want to compute the image $f_2$ that shows the object from a view different by the matrix $\mathsf V'\in\mathbb R^{4\times 4}$.

The 3D model typically contains the object in question, as well as its context, \eg, a car standing on a planar ground, which is very common for cars.
The particular type (triangular,  (NURBS) patches, procedural, \etc) and technical properties of the 3D model (face count, orientability, UV coordinates) are not relevant for our approach, as we only require that the model can be rendered into an image from a view.

First, let $p_1: \mathbb R^2\rightarrow\mathbb R^3$ be the \emph{position image}, which is the projection of $\mathcal S$ in the original view after hidden-surface removal.
We produce such images using deferred shading \cite{Deering1988} with $z$-buffering, \ie, generating positions instead of shaded colors in rendering.
Using deferred shading, we also compute $n_1: \mathbb R^2\rightarrow\mathbb R^3$, the \emph{normal image} computed from the derivative of $\mathcal S$; $r_1: \mathbb R^2\rightarrow\mathbb R^9$, the \emph{reflectance image}; and $L_1:\mathbb R^2\rightarrow\mathbb R^3$, the radiance (\refFig{Overview}).
Position, normal and reflectance images $p_2,n_2,r_2$ and $L_2$ from the novel view $\mathsf V'$ are produced in the same way.
Finally, we compute sampling quality and occlusion in a $z$-buffer-like test, formalized by a function \[
 q(\mathbf x) = \left\{
 \begin{array}{l l}
  \max(0, n(\mathbf x)\cdot v(\mathbf x))&
  \ \text{if\ } z(p(\mathbf x)-\mathsf V' p(\mathbf x)) >\epsilon\\
  0&
  \ \textrm{else,}
  \end{array}
  \right.
\]
where $v(\mathbf x)=\mathsf V'^{-1}p(\mathbf x)/||\mathsf V'^{-1}p(\mathbf x)||^2_2$ is the normalized viewing direction and $z$ returns the depth component of a vector.

Further we define a metric on all attributes as follows (\refFig{Weights}):
As a distance on positions $\Delta_\mathrm p$ we use Euclidean distance;
for normals, we use the dot product as the distance $\Delta_\mathrm n$;
for reflectance, we apply a perceptually linear Phong BRDF distance $\Delta_\mathrm r$ similar to \cite{Pellacini2000};
radiance and locality of two pixels $\Delta_\mathrm L$ and $\Delta_\mathrm l$ is again measured using  Euclidean distance.
\mycfigure{Weights}{
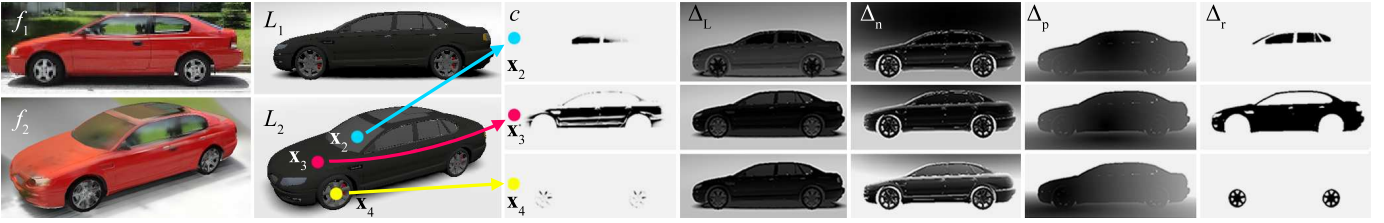
Reconstruction weights $c$ and distances $\Delta_{\{\mathrm L,\mathrm n,\mathrm p,\mathrm r\}}$ of three output locations $\mathbf x_{\{2,3,4\}}$ in respect to all positions from the input image.
}
Using all the above, we can now write the probability of assigning the appearance from pixel position $\mathbf x_1$ to the novel appearance at location $\mathbf x_2$ as \begin{align*}
	c(\mathbf x_1,\mathbf x_2)=
	q(\mathbf x_1)/
	(
	&w_\mathrm p \Delta_\mathrm p(p_1(\mathbf x_1), p_2(\mathbf x_2))+\\
	&w_\mathrm n\Delta_\mathrm n(n_1(\mathbf x_1), n_2(\mathbf x_2))+\\
	&w_\mathrm r\Delta_\mathrm r(r_1(\mathbf x_1), r_2(\mathbf x_2))+\\
	&w_\mathrm L\Delta_\mathrm L(L_1(\mathbf x_1), L_2(\mathbf x_2)))+\\
	&w_\mathrm l \Delta_\mathrm l(\mathbf x_1, \mathbf x_2)
	).
\end{align*}

\mysubsection{Reconstructing appearance}{ReconstructingAppearance}

Reconstructing a novel view is now a weighted sum over the input image.
The result image $f_2$ is reconstructed at location $\mathbf x_2$, as \begin{align}
f_2(\mathbf x_2)=
\left.
\int
c(\mathbf x_1,\mathbf x_2)^s
f_1(\mathbf x_1)
\mathrm d \mathbf x_1
\middle/
\int
c(\mathbf x_1,\mathbf x_2)^s
\mathrm d \mathbf x_1
\right.
,
\label{eq:ReconstructionAppear}
\end{align}
where $s$ is a sharpness parameter.
If $s$ is low, the reconstructed appearance is combined from many sources.
It is more reliable, but also more smooth.
If $s$ gets larger, fewer, but higher-quality observations dominate the solution.
For discrete images the integral above turns into a sum in practice.
Instead of aligning the 3D model $\mathcal S$ to the original image $f$ in 3D, we, for now, deform the original 2D image \cite{Beier1992} to align it to the images $p_1, n_1$ and $r_1$ of the 3D model $\mathcal S$.
\refSec{ViewSynthesis2} will discuss how to automatize this step.

We do not need to iterate over the entire image but only over a local neighborhood.
If $\mathbf x_2$ is the novel image pixel position, we run over a fixed-size neighborhood around location $\mathbf x_2+w(\mathbf x_2)$, where $w:\mathbb R^2\rightarrow\mathbb R^2$ is the flow between the novel view and the original view, \ie, where every pixel in the novel view is coming from in the original view.
This is because more correlated pixels are found in the neighborhood of the pixel position, that a world space position had in the original view.
It is produced as follows:
We again rasterize $\mathcal S$ from the view $\mathsf V'$ but store $w(\mathbf x)=\mathbf x-\rho(\mathsf V'p(\mathbf x))$ at the pixel with location $\mathbf x$, where $\rho: \mathbb R^3\rightarrow\mathbb R^2$ is the (perspective) projection from world to image space.
Flow is undefined for positions that were occluded in the original view.

We use a GPU to produce all guide images, parallel over output pixels, using deferred shading and to evaluate \refEq{ReconstructionAppear} on a 512$\times$512 image using a 128$\times$128 neighborhood in less than a second, allowing to interactively explore novel views by moving a virtual camera.

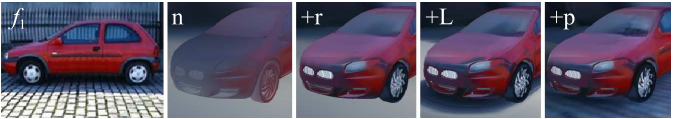
\myfigure{DistanceAnalysis}{Contribution of distances to the reconstruction weights \emph{(See text)}.}
For reconstruction, the weights $w$ are tuned by visual inspection of the result, which is easy thanks to the interactive feedback (\refFig{DistanceAnalysis}). The novel view can be synthesized as 2D image directly using \refEq{ReconstructionAppear}, but the reconstruction can also be expressed in terms of \textit{Reflectance Maps}~\cite{Horn1979} (or Lit Spheres~\cite{Sloan2001}). A Reflectance Map captures the orientation-dependent appearance of an object/object part. In other words, it is an association between surface normals and their color values. Here, a Reflectance Map can be computed by synthesizing the appearance on a sphere using \refEq{ReconstructionAppear}.

We now discuss the individual terms that contribute to the reconstruction:
First (\refFig{DistanceAnalysis},+$n$), if all weights except normal are zero, our approach is equivalent to the Lit Sphere model~\cite{Sloan2001}.
All details are missing here, except a global dependency on surface orientation.
Second (\refFig{DistanceAnalysis},+$r$), adding a reflectance term is equivalent to a Reflectance Map for each material.
This is equivalent to a ``continuous'' Reflectance Map depending on material parameters.
Third (\refFig{DistanceAnalysis},+$L$), the radiance term prefers appearance that is similar if multiple appearances are equivalent.
This is the case for the ground plane, that is either shadows or unshadowed in $L_1$ and $L_2$, which is transferred from $f_1$.
Finally (\refFig{DistanceAnalysis},+$p$), adding a dependency on position prefers local appearance if everything else is similar.
In our example, the cobblestone pattern that is not visible in any feature, but correlates with position is reproduced.
Note, that when adding one component, no other component is degraded.

The 3D model is fit into the unit cube to normalize positions and make them comparable for different scenes.
The settings in \refFig{DistanceAnalysis} are
$w_\mathrm n=1$,
$w_\mathrm r=2$,
$w_\mathrm L=0.1$,
$w_\mathrm p=0.01$,
$w_\mathrm l=1$
.
The sharpness is set to $s=3$.
We keep those constant for all results reported throughout the section. A visualization of the sampling in the directional domain is seen in \refFig{LitSpheres}.
\myfigure{LitSpheres}{
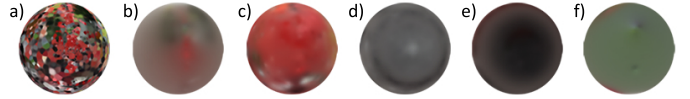
Reflectance Map for the input image $f_1$ seen in \protect\refFig{DistanceAnalysis}:
\emph{a):} Reconstructed using nearest-neighbor.
Note the increasing density towards frontal views.
\emph{b):} Our reconstruction without position, material and radiance information (only normal differences).
\emph{c--f):} Our reconstruction with all information included for different materials of the object (body, rim, tire and  windshield).
}


Regarding illumination, the reference images are illuminated using one ``representative`` lighting (skylight with ambient occlusion), that can be assumed
to also be present in the 2D image for cars. For background, we use a large white background sphere for all 3D models such that the function p, n, r, L, $\etc$
are always defined. When evaluating Eq.~\ref{eq:ReconstructionAppear} on background, distance to foreground is so large that it does not contribute, so similarity is based on 3D position and normal alone.
Effectively the background is projected onto a sphere, including proper occlusion.

\mycfigure{Overview3}{
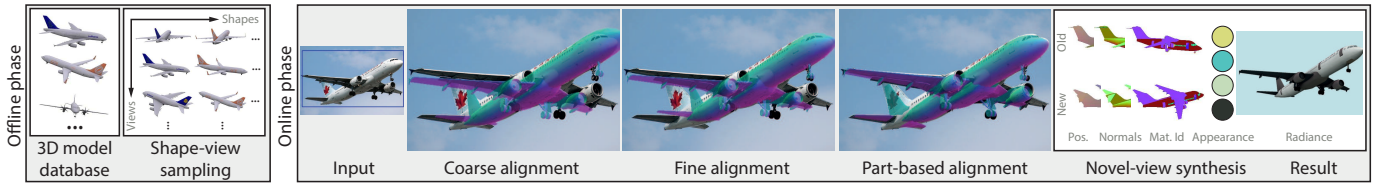
Overview of our approach comprising an off-line and an on-line phase.
The off-line phase samples shapes from a database in many views.
In the on-line phase, coarse, fine and per-part 2D-to-3D alignment is performed which is based on a coarse image segmentation derived from user input (scribbles).
Finally, novel views are synthesized by rendering the 3D shape in the old and novel views and transferring view-dependent appearance between those images.
}

\mysection{2D-to-3D Alignment}{Alignment}

The novel view synthesis method from the previous section assumes that the 3D model and the 2D object are well aligned. In this section we present an approach to 2D-image-to-3D-model-alignment that requires minimal user interaction.

The alignment component comprises an off-line and an on-line phase.
In the off-line phase (\refSec{OfflineAlignment}), a database of 3D shapes is analyzed to learn a set of discriminative shape-view detectors.
There is one discrete detector for each 3D shape in each view.
To this end all shapes are rendered from all views using different illuminations.
Learning these detectors can take several hours, but is only conducted once.
Besides luminance, our detectors also use coarse segmentations. Also in the offline phase, 3D shapes are segmented into parts.

The on-line phase (\refSec{OnlineAlignment}) proceeds in a coarse-to-fine and discrete-to-continuous scheme.
First, the discrete 3D shape and view for the query 2D image are found using the previously learned detectors.
Detectors account both for luminance and a segmentation produced from user input.
Having found the best discrete view however, will not result in a precise alignment.
To this end, the view is refined continuously in the next step.
Alternative views of the best 3D shape are rendered on-the-fly to find the view resulting in the best match to the query 2D image.
Finally, as the closest 3D shape in the database is often not identical to the one in the query 2D image due to within class variability in the considered object categories, a per-part alignment is found that continuously refines the 3D shape.

Producing a physically or functionally meaningful 3D shape from such part-based alignment would require an optimization that reliably accounts for physical and functional part relations for all objects in an entire 3D shape database.
Instead of addressing this daunting task, we advocate to use this part-based alignment only for improved localization when capturing view-dependent appearance for the novel view-synthesis (\refSec{ViewSynthesis2}).
For each material of the 3D shape, directional-dependent appearance is sampled using the normal on all image areas with parts of that material.
This allows to re-render the shape from novel views with fine spatial details and faithful view-dependent appearance in real-time, as we show in~\refSec{ViewSynthesis2}.

\mysubsection{Off-line phase}{OfflineAlignment}
Input to the off-line phase is a database of 3D shapes.
First, we render 2D images from these shapes by randomly sampling view parameters. 
We call these samples ``shape-view samples'' (\refFig{Sampling}).
For each sample we train an exemplar-based discriminative shape-view detector. 
In order to facilitate the part alignment in the last stage, we also segment the mesh into parts that allow for a fine-granular alignment accounting for the variations observed in the instances at runtime.

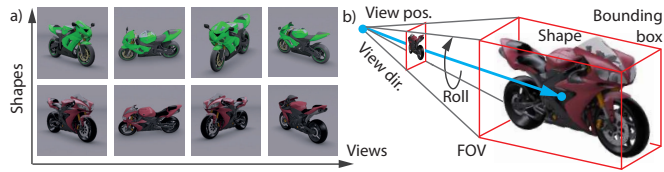
\myfigure{Sampling}{Sampling of the shape-view-space \emph{(a)} and sampling of a view for a specific shape \emph{(b)}.}

\paragraph*{3D Shapes}
Our method is based on a database of 3D shapes. We do not require any specific (\eg, meshing) fidelity as long as the models can be rendered from many views.
Therefore our approach is able to tap into rich on-line resources, which provide a vast amount of 3D shapes, capturing many shape classes with their respective intra-class variation.
For our experiments, we use 3D models from 3D Warehouse, as they are indexed in\cite{Chang2015}.
Our current database consists of 10 object classes ({\sc airliner, fighter jet, bicycle, double bus, folding chair, office chair, station wagon, sedan, sport motorbike, sofa}) that are on average represented by 50-250 instances for each class.

\paragraph*{Sampling}
We sample from the seven-dimensional space of all possible views, that is, all view positions (3D), view orientations (2D) camera roll angle (1D) and field of views (FOV, 1D) as shown in \refFig{Sampling}(b).
In order to reasonably bound the choice of views at training time, the positions are selected from a shell between the bounding sphere and a sphere twice as large.
The orientation is selected by picking a random point inside the bounding sphere.
Roll is picked randomly uniform.
The FOV is randomly chosen corresponding to the range of focal length between 20 and 200\,mm.
Finally, the distance along the view direction is set so that the complete shape fits the viewport.

\paragraph*{Rendering}
To produce a 2D image for each shape-view sample, we use image-based lighting \cite{Debevec1998}, from an environment map (``Uffizi'' by Debevec~\etal~\cite{Debevec1998}) at the default orientation in proximity of the camera in combination with ambient occlusion \cite{Zhukov1998}.
In total, we produce on average $n_\mathrm s=60,000$ images per class, each in a resolution of 512$\times$512 with 2$\times$2 supersampling.
Transparent materials are rendered as such, but without reflections or refractions.
After rendering, every image is converted into a Histogram-of-Oriented-Gradients (HOG)\footnote{Our approach can also be used with different features (\eg from pretrained CNNs). We choose HOG because of their proven ability to match different modalities \cite{Aubry2013, Malisiewicz2011, Aubry2014, shrivastava2011}.} \cite{hog} image representation based on the contained object's bounding box.
The spatial sampling of the HOG cells is chosen to result in approximately 150 cells -- corresponding to a 1350-dimensional feature vector.
Additional to rendering luminance, a binary foreground mask of the object is rendered and stored in a HOG representation in the same way.

\paragraph*{Discriminative shape-view detectors}
The construction of shape-view detectors is based on the global mask and luminance templates.
For each template, a classifier is computed based on Linear Discriminant Analysis (LDA)~\cite{Hariharan2012}.
The classification score between a template's HOG representation $\mathbf d$ and a test HOG representation $\mathbf x$ is defined as
$
S'(\mathbf d,\mathbf x) =  (\Sigma^{-1} (\mathbf d - \mathbf \mu))^\mathsf T \mathbf x,
$
with $\Sigma$ and $\mu$ being the covariance and mean of the negative data respectively.
These were computed using a set of 100,000 random patches extracted from natural images. For the mask negative statistics we use the segmentation masks from the Pascal~VOC~2012~\cite{Everingham2012} images.
Since the HOG dimension of the templates varies, it is intractable to calculate $\Sigma$ and $\mu$ for all possible HOG sizes.
Therefore $\Sigma$ is estimated using the auto-covariance matrix $\Gamma$ as done by Hariharan~\etal\cite{Hariharan2012}.
To make the responses of the different exemplar classifiers comparable, the affine calibration approach of Aubry~\etal\cite{Aubry2014} is used based on negative data, producing a calibrated classifier $S=\alpha S'+\beta$.

\paragraph*{Part segmentation}
To improve novel view-synthesis quality, we propose per-part alignment (\refSec{PartAlignment}).
To this end, all input 3D shapes need to be segmented into parts.
Segmenting a mesh into meaningful parts is a difficult problem that we try to avoid. Instead, we advocate for a simple mesh segmentation approach.
Initially, the mesh is segmented into connected components based on triangle connectivity and material index. We then iteratively merge the part with the smallest area into one of its neighbor parts until a class-specific number of parts (typically eight) remains.

\mysubsection{On-line phase}{OnlineAlignment}
Input to the on-line phase is a 2D \emph{query} image and output is a 3D shape, a view matrix and a per-part model transformation matrix.
The on-line phase has three steps:
a) coarse and discrete shape-view alignment by detection,
b) fine-grained alignment to continuously refine the view by producing examples on-the-fly and c)
part-based alignment that independently and continuously moves individual 3D shape parts. The effect of each step with respect to the 2D-to-3D alignment is shown in \refFig{AlignmentEffect}

\myfigure{AlignmentEffect}{
Effect of alignment steps on appearance and geometry matching:
The \emph{coarse} match is the closest exemplar of examples and fails to align even in a rigid sense.
\emph{Fine} alignment finds a good rigid alignment, but still cannot produce a match as the 3D shape is different from the one in the database.
Only the \emph{part-based} alignment puts the arm rests and wheels in the right location from where appearance can be transferred.}

We assume the input image to be segmented into foreground and background, either using an automated approach (\eg semantic image segmentation \cite{long2014fully}) or a sufficient amount of user interaction \cite{rother2004grabcut}.
User interaction is required to indicate where the object to be lifted is located in the query image anyway, for example when multiple objects are present in the query image.
A typical way of selection would be to draw a rough 2D bounding box.
Our results always show the bounding box or scribbles that were applied by the user on the images for lifting to 3D. In all results, the user input is minimal and on the order of a few seconds.

\mysubsubsection{Coarse alignment}{CoarseAlignment}
We approach coarse alignment of a 3D models to an 2D input image with a mask by retrieving the best matching discrete shape and view using the previously trained shape-view-detectors. Our proposed speed-ups are key to incorporating this challenging sub-task in an interactive system.

\paragraph*{Detection}
First, the input image and its mask are converted to a HOG representation $\mathbf x$ and $\mathbf x_\mathrm{mask}$ respectively.
Next, the alignment score is the combination with equal weights of the scores $S(\mathbf d,\mathbf x)$ and $S(\mathbf d_\mathrm{mask},\mathbf x_\mathrm{mask})$ of all samples $\mathbf d$ and $\mathbf d_\mathrm{mask}$.
The sample with the best score is further considered in the processing. Using a single global HOG representation is faster than \eg the discriminative part based approach advocated by \cite{Aubry2014}.

\paragraph*{Pruning}
Pruning is used to accelerate detection.
The number of shape-view samples becomes large when a finer alignment is desired, different FOVs are used and the number of shapes is increased.
We found the segmentation mask created from the user click, stroke or bounding box to be an excellent means of culling many samples with very little effort.
To this end, samples with bounding boxes that are different in size by more than one HOG cell from the query bounding box are excluded.
Boxes of a certain range of sizes can be found in time complexity sub-linear to the number of samples, typically eliminating a large part of the samples early-on.

For instance, in our simplest setting where only 128 views per 3D model are considered, on average only 633 out of 6400 samples per class need to be tested.
The pruning alone is 1 order of magnitude faster than actually computing a detection response.
We conclude that our pruning is an effective means to reduce computational effort for 2D-to-3D alignment.

\mysubsubsection{Fine alignment}{FineAlignment}

The fine alignment starts from the coarse alignment and refines the view.
Furthermore, as our database shapes do not always match the ones in the input image, we add more variations to our 3D model by allowing non-uniform scale along the $x$, $y$ and $z-$axes.
To this end, we parameterize the continuous 10-dimensional space, that is, rotations (3D), translations (3D), FOV (1D) and model scale (3D).
The best state that minimizes a cost function is found using simulated annealing.
Luminance and mask HOG distance to the 2D query image are used as a cost function with equal weights, without any training being performed.
In each iteration, the current state is changed to an alternative state by modifying 3 random components (out of 10) of the current state. An alternative view of the 3D shape is then rendered into a 2D image and the cost function is computed. An alternative state that decreases the current cost function is always kept while state resulting in an increase is kept according to a probability based on the Boltzmann factor \cite{Kirkpatrick1983}.
With progressing iterations, the alternative state differs less from the current state.
Results of this step are a refined view and project matrices, and a scaled 3D model.
300 iterations are used for all the results shown in the paper.

\mysubsubsection{Per-part alignment}{PartAlignment}
In this final step, the view is held fixed and the individual parts of the 3D shape are transformed independently.
The refinement uses multiple passes of simulated annealing.
In each pass, a random part is chosen and simulated annealing (in the spirit of fine alignment step) is used.
For each part, we allow rotation (3D), translation (3D) and model scale (3D).
We further clamp these parameters to a threshold so that only small perturbations from fine alignment are allowed.
Again, the mask and luminance HOG distance is used to either keep or discard the part transformation.
We use 20 passes, each pass with 20 iterations for all results shown in the paper.

The outcome is not a consistent 3D shape, as part deformation neglects physical or functional part relationships.
However, our novel view-synthesis (\refSec{ViewSynthesis2}) only requires individual parts to be well-placed absolutely to capture appearance while their relative placement is not relevant.

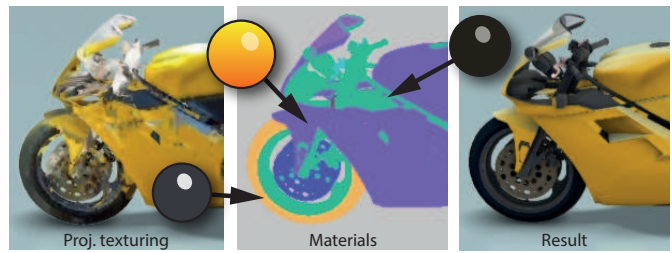
\myfigure{NovelViewSynthesis}{Novel views synthesis: For every material index \emph{(2nd col.)}, a Reflectance Map is constructed based on the appearance of the object in the image \emph{(1st col.)} and the normals in view space. These appearance maps are then used to synthesize appearance for a novel view \emph{(3rd col.)}.}

\mysection{Novel-view Synthesis Revisited}{ViewSynthesis2}
In \refSec{NovelViewSynthesis} we presented a method to synthesize novel views of an object with the guidance of a 3D model. While the obtained results have high level of realism, the synthesis is view-independent. This means that when we change viewpoint in 3D, specific visual elements that are view dependent such as specularity remains unaffected. However, in a scenario that involves user interaction and manipulation of the object in 3D, a view dependent synthesis is required. For this reason we extend the synthesis method of \refSec{NovelViewSynthesis} to capture view specific information.

\mycfigure{materialRMs}{
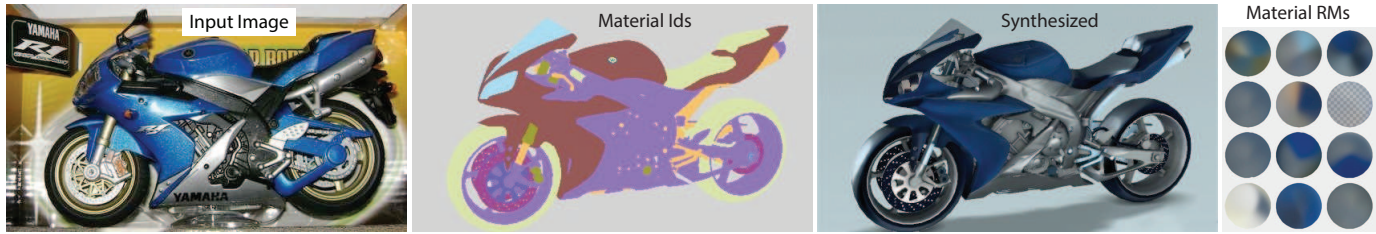
The improved novel-view synthesis (\protect\refSec{ViewSynthesis2}) segments the input image into parts using the material IDs of a 3D mesh as a guide.
To synthesize a new image, a reflectance map is extracted for every material -- shown on the left -- , allowing high speed and view-dependent appearance.
}

The 2D-to-3D alignment from the previous section identified for an image $f_1$ the shape $\mathcal S$ (a polygonal mesh), a finely-aligned view $\mathsf V$ (a camera transformation) and a model matrix $\mathsf M_j$ for each part $j$.
We now would like to synthesize a 2D image from a novel view $\mathsf V'$ in real-time, with detailed appearance also in the parts that were unobserved in the query image.
In particular, we would like the appearance to be view-dependent, \ie, a highlight on a car body should not appear as if it was painted, but move in a faithful way to convey the material and visual essence of a car.

To this end, we define the RGB appearance $A_i(\omega)\in\Omega^+\rightarrow\mathbb R^3$ of the $i$-th material from direction $\omega$ in view space as \begin{align}
A_i(\omega)&=
\int
\gamma_i(\omega, \mathbf x)
f_1(\mathbf x)
\mathrm d x
/
\int
\gamma_i(\omega, \mathbf x)
\mathrm d x,
\end{align}
where
$f_1(\mathbf x)$ is the RGB image appearance at 2D position $\mathbf x\in(0,1)^2$ and
$\gamma_i(\omega, \mathbf x)$ is a weighting kernel
\begin{align}
\gamma_i(\omega, \mathbf x)&=
q_i(\mathbf x)
\exp(-\alpha|1-\left<\mathsf R n(\mathbf x),\omega\right>|^2),
\end{align}
where
$q_i(\mathbf x)$ is a compatibility function that is 0 if the image at position $\mathbf x$ does not belong to the mask, otherwise 1 if material index at $\mathbf x$ is $i$ and finally, 0.0001 if material at $\mathbf x$ is not $i$. The last weight is used to enforce colors for materials that are not observed at any $\mathbf x$ in $f_1$. $n(\mathbf x)$ is the world space-normal,
$\mathsf R$ is the upper-left $3\times 3$ sub-matrix of the inverse transpose of the $4\times 4$ matrix $\mathsf V$ and
$\alpha$ is a sharpness parameter, typically set to $\alpha=4$.
The novel image from $\mathsf V'$ is defined as
\begin{align}
f_2(\mathbf x) = A'_i(n(\mathbf x)) = A_i(\mathsf R' n(\mathbf x)),
\end{align}
 for material $i$ and the rotational part $\mathsf R'$ of $\mathsf V'$.

In the discrete case, the function $A_i$ becomes a look-up table of directions (we use $32\times 32$ directions). This table is essentially a Reflectance Map~\cite{Horn1979}. 

Each table entry is computed as a weighted sum over all input image pixels and based on the normal and material id of $\mathcal S$ rendered from $\mathsf V$ using a per-part model matrix $\mathsf M_j$.
\refFig{NovelViewSynthesis} illustrates the concept. By the aligned 3D geometry we gather data off the 2D image in order to estimate Reflectance Map models that are used to render novel views. The available material information together allows for a much crisper rendition and truthful interaction of light with the material in contrast to projective texturing.
As $A_i(\omega)$ is a convex combination of input pixels, we noticed a loss in contrast.
As a solution, we perform CIE LAB histogram matching  between the smoothed  histogram of $A'$ and the histogram of the input image $f_1$.

At render time, a single material-dependent Reflectance Map is required (\refFig{materialRMs}), allowing to produce a novel view with several hundred frames per second in high resolutions, in our images combined with ambient occlusion \cite{Zhukov1998}.

Regarding the Reflectance Maps, there are two main differences compared to~\refSec{NovelViewSynthesis}. First, in this section the goal is to estimate directly the entries of the Reflectance Map, while previously we were estimating the appearance for every disoccluded pixel in the image. In addition, here we consider viewer-centered normals. Therefore, a change of viewpoint affects the location of specularities or artifacts. In~\refSec{NovelViewSynthesis} such visual elements were considered as "painted" on the object.


\mysection{Experiments}{Experiments}
In this section we present qualitative and quantitative results for the novel view synthesis and 2D-to-3D alignment methods presented in this paper.

\mysubsection{Synthetic Viewpoint Dataset}{SyntheticDataset}

We apply our novel view synthesis approach from \refSec{NovelViewSynthesis} in order to generate new views for a set of 26 sideview images of cars from PASCAL VOC 2007~\cite{PascalVOC2007}. Here, we manually aligned the images with the 3D models (about 2\,h effort) and we generate for each image 9 synthetic views by sampling the viewing sphere.
Examples of the synthesized data are shown in \refFig{NovelViews}. Note how effects of global illumination on the vehicle as well as shadow below the car are preserved in many images. The disocclusion areas are filled in in a plausible and natural way. The transition between the visible and ``hallucinated'' part is seamless.
\myfigure{NovelViews}{
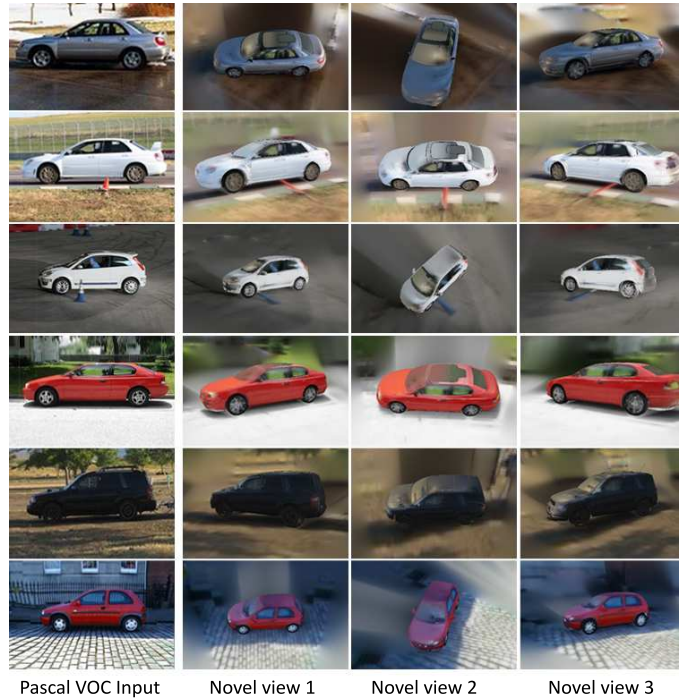
Startting from PASCAL VOC input training 2D images \emph{(1st column)} we produce arbitrary novel-view training images \emph{(other  columns)}.
}

\mysubsection{Training data augmentation}{DataAugmentation}
Object detection and classification approaches have seen substantial improvements over the last decade. One driving factor is the availability of training data that is representative for the test scenarios of interest.
However, the construction of such data sets is tedious and yet does not capture all aspects of variability in the classes that it contains. In particular the sampling of untypical examples or viewpoints is often lacking \cite{HoiemECCV12}.
More specifically, the popular PASCAL VOC benchmark  \cite{PascalVOC2007} provides a good sampling of intra-class variations but without exhaustive view-point sampling. Other data sets like EPFL Cars \cite{OzuysalLF09} provide dense view sampling (without azimuth) but do not capture intra-class variation well.

Next we use the aforementioned Synthetic Viewpoint Dataset to ``augment'' the PASCAL VOC training data in order to represent the intra-class variation together with a better viewpoint sampling that includes atypical views. Our study focuses on the ``car'' class.

Given this augmented dataset we run a series of experiment to underline the validity of our approach.
In all experiments, we use the state-of-the-art Deformable Part Model (DPM version 5)~\cite{dpmPAMI10}.

\begin{table}[h]
\centering
\caption{Performance of the DPM detector when trained in different data and with different number of components ($N$).}
\vspace{-0.3cm}
\label{tab:sideviewComparison}
\begin{tabular}{rcccc}
\toprule
Data & \multicolumn{2}{c}{Avg. precision (\%)} &  \multicolumn{2}{c}{\#Views}  \\
 & $N=1$		&$N=3$		& Real & Synth.\\
\midrule
Side          & 15.4   &16.2  & 26 &  -   \\
Rendered	& 11.5   &12.7  & -     & 26 \\
Synthetic  	& 15.0   &14.5  &  -    & 26 \\
\bottomrule
\end{tabular}
\end{table}
\paragraph*{Pilot study -- Re-synthesis}
As an initial test, we investigate how much the synthesized views affect the performance, compared with the real images and direct  renderings of the 3D models.
We perform this study on the 26 sideviews which are resynthesized using the method from \refSec{ReconstructingAppearance} by treating the visible part of the car as a disocclusion.
Using this data we train a DPM detector and we test on the whole VOC test set.
Table~\ref{tab:sideviewComparison} shows the performance numbers in average precision (PASCAL VOC standard criterion) for different number of components (columns) in the DPM.
The first line represents training on the real 26 sideviews.
Then we repeat the training but in this case we have removed the car using the ``Context Aware''-tool from Adobe Photoshop and we replaced it with a rendering of the 3D model that corresponds to that type of car (Sedan, SUV or Compact) (\refFig{pilotStudy}).
We observe that the performance is much lower, due to the lack of variation in the car appearance.
The last row  corresponds to training on the 26 resynthesized cars using our approach. The performance is very close to the training on the real cars which provides first evidence that our method is indeed able to generate the kind of realism that is needed to successfully train an object detection algorithm without a strong loss in performance that is often observed in such settings.
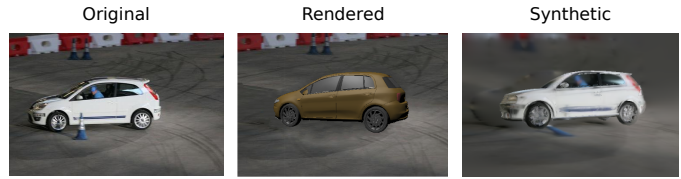
\myfigure{pilotStudy}{Pilot study training data: a) original image, b) rendered 3D model, c) our synthesized object.}

\begin{table}[b]
\centering
\caption{Performance of the DPM detector on PASCAL VOC dataset with varying training set and  different number of components ($N$). Evaluation is performed on the full test set as well as a subset of rare viewpoints.}
\label{tbl:VOCFullTable}
\vspace{0.1cm}
\begin{tabular}{llccccc}
\toprule
Test & Train &  \multicolumn{3}{c}{Avg. precision (\%)} & \multicolumn{2}{c}{\#Views}\\
\midrule
&			&$N=3$		&$N=4$		&$N=5$& Real& Synth. \\
\midrule
\multirow{4}{*}[4pt]{ \begin{turn}{-90}VOC 2007\end{turn} }
&Side			&16.2	&18.4  &16.7 & 26 & -\\
&Side+synth		&30.2	&31.4  &33.2 & 26 & 728\\
&Full			&51.7	&53.4  &50.7 & 1250 & -\\
&Full+synth		&50.2	&53.1  &50.9 & 1250 & 728\\

\midrule

\multirow{4}{*}[2pt]{ \begin{turn}{-90}VOC rare\end{turn} }
&Side			&11.9	&11.6  &10.3 & 26 & -\\
&Side+synth		&23.2	&30.2  &32.9 & 26 & 728\\
&Full			&51.9	&52.5  &51.8 & 1250 & -\\
&Full+synth	&55.0	&57.3  &53.1 & 1250 & 728\\

\bottomrule
\end{tabular}
\end{table}

\paragraph*{PASCAL VOC data set}
We continue by using our whole synthetic viewpoint dataset and mixing it with different portions of real data.
Quantitative results varying the number of components in the DPM are shown in \refTbl{VOCFullTable} and performance plots in \refFig{VOCPlots}. We first focus on the upper half of Table~\ref{tbl:VOCFullTable} where we test on the standard PASCAL VOC test set.
We start by training only on the 26 sideviews (Side) that we consider for augmentation and compare the performance to a model on the same 26 views amplified by 728 views synthesized by our approach (Side+synth). We observe improvements between  $14$ to $16.5\%$ by adding our synthesized views. We now compare the model trained on the full VOC training set (Full) to a version to which we add our synthetic views. In this setting we do not  observe an improvement in average precision, but rather comparable performance.
However, if we inspect the associated precision recall curve in \refFig{VOCPlots} (left) more closely, we observe that our model improves in the high precision regime. It produces no single false positive until $15\%$ of the positives are detected (recall).

In order to generate further insights into our model, we perform a study similar to the one proposed in \cite{HoiemECCV12}. Here we focus on rare viewpoints of the cars by selecting a subset of the PASCAL VOC testset where the top of the car becomes visible. Our reasoning is that those cases are difficult for the standard model as this part of the viewpoint distribution is poorly sampled.
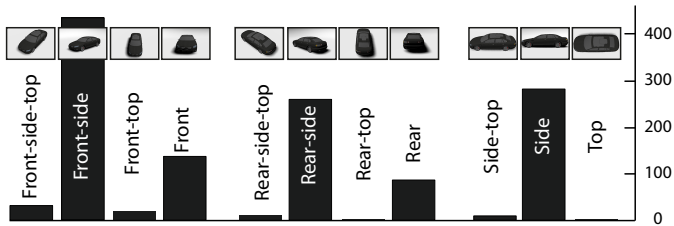
\myfigure{viewpointsCar}{Object side visibility for car VOC2007 test data set.}
\refFig{viewpointsCar} confirms out intuition about the view distribution in the VOC dataset. The statistic shows that all views involving the visibility of the top are under-represented.
The lower part of \refTbl{VOCFullTable} performs an analysis for detection of such rare viewpoints. We see strong improvements on those rare viewpoints of up to $22.6\%$ for training from sideviews and up to $4.8\%$ for the full training set.

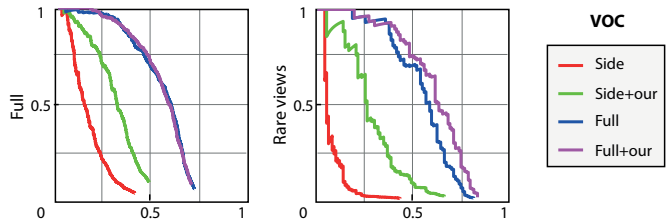
\myfigure{VOCPlots}{Performance (Average-Precision) of the ``side'' and ``full'' versions of PASCAL VOC data set with and without novel views produced using our approach. Evaluation is performed on the full PASCAL VOC test set (left) as well as a subset of rare viewpoints (right).}

\paragraph*{UCLA data set}
In order to get a more realistic estimate of the performance across viewpoints we turn to the UCLA cars data set \cite{Hu12} which has been designed with a more uniform viewpoint sampling in mind.
The test set consists of 200 images that cover better the viewing sphere.
\begin{table}[h]
\centering
\caption{Performance of the DPM detector on UCLA dataset with varying training set (images from VOC) and  different number of components ($N$).Evaluation is performed on the full test set as well as a subset of rare viewpoints.}
\label{tab:ucla}
\vspace{0.1cm}
\begin{tabular}{llccccc}
\toprule
Test & Train &  \multicolumn{3}{c}{Avg. precision (\%)} & \multicolumn{2}{c}{\#Views}\\
\midrule
&			&$N=3$		&$N=4$		&$N=5$& Real& Nov. \\
\midrule
\multirow{4}{*}[4pt]{ \begin{turn}{-90}UCLA\end{turn} }
&Side			&41.5	&43.5  &41.1 & 26 & -\\
&Side+synth		&83.0	&82.4  &85.4 	& 26 & 728\\
&Full			&75.4	&78.7  &78.3 	& 1250 & -\\
&Full+synth		&84.0	&86.0  &85.2	& 1250 & 728\\

\midrule

\multirow{4}{*}[8pt]{ \begin{turn}{-90}UCLA rare\end{turn} }
&Side			&40.5	&44.2  &39.7 & 26 & -\\
&Side+synth		&82.2	&81.9  &85.1 & 26 & 728\\
&Full			&69.8	&73.7  &72.5 & 1250 & -\\
&Full+synth		&81.4	&83.6  &83.3 & 1250 & 728\\
\bottomrule
\end{tabular}
\end{table}

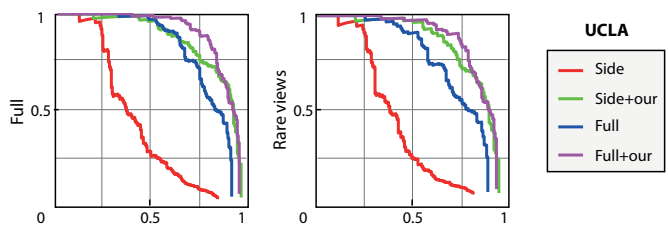
\myfigure{UCLAPlots}{Performance (Average-Precision) of the ``side'' and ``full'' versions of training  set (VOC) with and without novel views synthesized by approach. Evaluation is performed on the full UCLA test set (left) as well as a subset of rare viewpoints (right).}

In Tab.\ \ref{tab:ucla} and \refFig{UCLAPlots} we show the performance plots of the DPM detectors trained on the same dataset as in the previous section.
For the full UCLA test set (upper half of Table~\ref{tab:ucla}) we observe drastic improvements by adding our synthesized data. For the sideviews the improvement is between $38.9$ and $44.3\%$ and for the full set between $6.9$ and $9.6\%$. Remarkably, our detector trained on the amplified 26 sidesviews (Side+synth) outperforms the model trained on the full PASCAL dataset (Full) which had access to 48 times more real training examples (26 sideviews vs. 1250 real examples). The improvements can be seen more clearly in \refFig{UCLAPlots}. We also provide the rare viewpoint analysis for this dataset. Here the improvements are even more pronounced. From the precision recall curve we see that the model trained on the amplified sideviews (Side+synth) is outperforming the model trained on the full set. In this case 26 real training examples in combination with our augmentation method is enough to get close to the best performance on this data (Full+ours).

\mysubsection{2D-to-3D alignment}{2Dto3DResults}
\mycfigure{AlignError}{
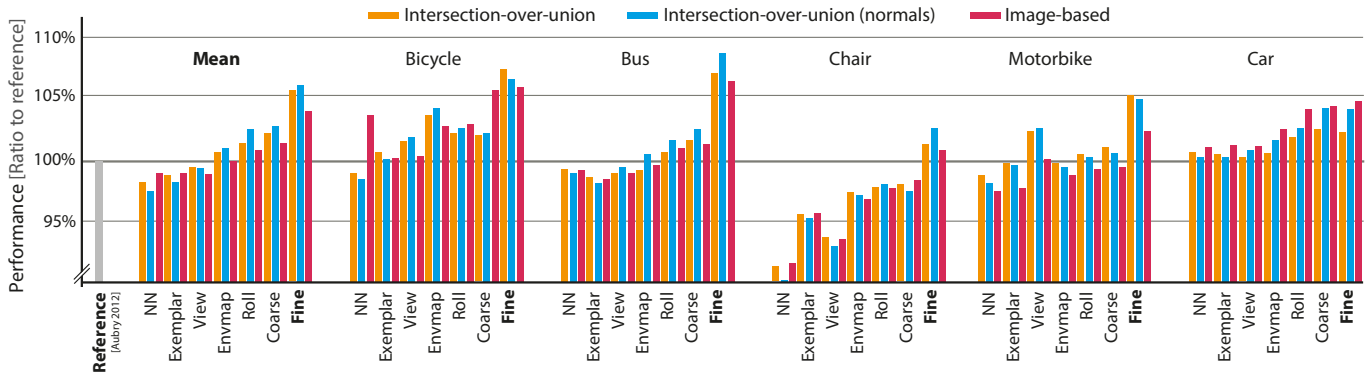
Comparing performance of several variants of our approach and a baseline \protect\cite{Aubry2013}.
Performance is shown as vertical bars in units of ratio to the baseline (more is better).
A group of bars is the outcome on one dataset, where the first one is the mean of all the others.
Inside each group, every triplet of bars is a varinat of our algorithm: nearest neighbor (NN); exemplar SVM, the same with increasing variation (view, envmap , roll) and finally the our full method, including fine alignment.
Finally, we compare performance using different metrics encoded as different colors: A classic intersection-over-union, our extenstion using normals and an image-based metric that assess the image similarity to the final image produced.
}

This section evaluates our approach in terms of
 quantitative performance according to different metrics and computational speed, a user study and visual quality of the renderings.

\mysubsubsection{Quantitative results}{QuantitativeResults}

We compare our 2D-to-3D alignment approach to previous methods in terms of different metrics.
As previous methods we consider eight alternatives:
the method of Aubry~\etal\cite{Aubry2013},
a method using light field descriptors \cite{Chen2003},
a method using nearest neighbor and
five simplified variants of our approach.
As metrics we use
intersection-over-union (IoU) of segmentation masks,
normal-aware IoU on the same,
and a novel metric using image differences of novel-view rendering.

\paragraph*{Methods}
The method by Aubry~\etal\cite{Aubry2013} uses discriminative patches on luminance.
We consider it our reference and would like to achieve similar or better performance, but in less time.
To quantify the effect of discriminative learning, we include nearest-neighbor matching in HOG space of luminance (NN).
Light field descriptors \cite{Chen2003} (LFD) are a popular approach in computer graphics \cite{Xu2011} thanks to their simplicity (they are based only on the silhouette of the object).
We would like to know if adding silhouette information to the luminance, as well as discriminative learning, improves the alignment performance.
Additionally we consider several increments in sampling variation for our approach:
a template LDA classifier trained on the same rendered images as Aubry (Exemplar),
adding viewpoint samples (View),
additional samples with environment map rotation (Envmap),
additional samples with camera roll (Roll)
and finally the two steps of
our full approach (Coarse and Fine) that include mask information. In each of these steps we increase the number of training images by including a certain type of variation in the rendering process (\eg denser viewpoint sampling, different lightning directions, \etc). These types of variation are expected to appear in the real images during testing.

\paragraph*{Metrics}
Intersection-over-union is a classic measure of alignment quality but remains agnostic to what is aligned.
Therefore, we also consider a normal-aware Intersection-over-union metric extension to this metric where agreeing pixels are not counted as 1, but as the dot product of the underlying screen space normals.
Doing so, only surfaces with a similar orientation are counted as a success, which is essential to the targeted synthesis approach.
The main objective of our approach is to produce an image of the object in the input from a novel view.
Therefore, we suggest to also use image differences between synthesized novel view-images as a third metric.
For two 2D-to-3D alignments A (ground truth) and B (the other), it is defined as the image distance
(\eg $1-$ the normalized $L_2$ difference in rgb values) between a novel view-synthesis result image when using A and when using B, divided by the union of the pixels that the novel views cover.
To produce the novel views, we apply projective texturing to both A and B and then rotate them $10^\circ$ in the $y$ axis.
This metric directly measure the effect of alignment errors on the synthesized images.

\myfigure{TimeError}{
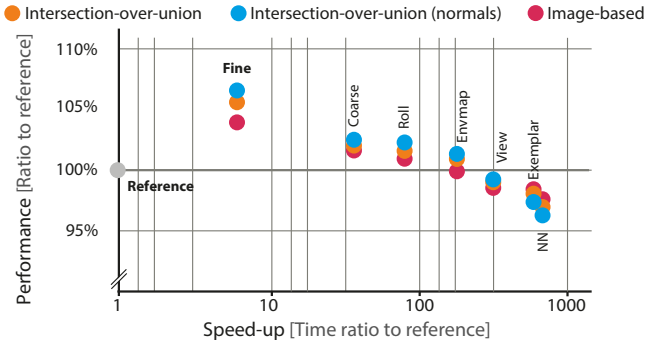
Relation of performance  shown in \protect\refFig{AlignError} to speedup.
The vertical axis is in the same units as \protect\refFig{AlignError}: ratio to a references according to different metrics encodet as colors.
The horizontal axis shows the speedup-ratio to the reference in log scale.
The reference is shown in grey with a speedup and performance ratio of one.
We see that our full method is both faster and better.
}

\paragraph*{Results}
2D-to-3D alignment was performed on 80 random images\footnote{All test images and results are available in the project website.} from 5 subclasses of Pascal3D~\cite{xiang2014} (Imagenet subset), namely station wagon, sport motorbike, folding chair, double bus, mountain bicycle, using the nine methods just described and the result was evaluated according to each of the three metric introduced. As ground truth we use the annotation from Pascal3D, even if it is noisy in many cases.
Performance is stated as the ratio to the outcome of the reference method (Aubry~\etal~\cite{Aubry2014}), \eg performance identical to the reference would be 100\,\%.
Using the ratio, all measures we propose become comparable (encoded as colors in \refFig{AlignError} and \refFig{TimeError}).
Performance is stated as the mean over classes, as well as for the five individual classes. While we have also conducted experiments using light field descriptors \cite{zhang2002integrated}, we leave them out of the plots as they have shown very poor performance of approximately $80\%$ on average consistently across all metrics.

In \refFig{AlignError} we present the effect of including complementary types of variation in the training data for 5 different classes. We show that by adding the aforementioned sampling schemes we improve the performance \wrt to the reference consistently for all metrics. Note that Exemplar performs better than NN for the same training data, similar to what is reported in \cite{shrivastava2011, Malisiewicz2011}, thus we show the comparison for the initial setting only.

Moreover, the results in \refFig{TimeError} show that our accelerated alignment procedure reaches speed-ups up to a factor of $\approx$ 600 with only moderate loss in performance (\eg ``exemplar''). This allows us to consecutively add the more elaborate sampling schemes without sacrificing the user-friendliness in terms of time. Our most complex sampling scheme -- constituting our final coarse alignment procedure (``coarse'') -- achieves an improvement of 2-3\% while still maintaining an $\approx$ 40 fold speedup. Our fine-alignment stage with on-the-fly rendering gives us on average a 5\% improvement, while still being roughly 5 times faster. The results are consistent across the 3 metrics that we have investigated.

Regarding the absolute computation time of our framework, most of it is spend in fine alignment. On average, 0.45\,s are spend for coarse, 2.4\,s for fine, and 2.1\,s for per part alignment.


\paragraph*{User study}
We asked human subjects to compare our fine alignment result to the exemplar result in order to assess the alignment quality.
The experiment comprised of 50 trials, where in each, participants were shown the result of both alignments for one out of 10 random instances from 5 classes of Imagenet ({\sc station wagon, sport motorbike, folding chair, double bus, mountain bicycle}).
Alignment was shown as two images in a random horizontal layout where in each the 3D shape is rendered using the normal shading and composed on top of the query image, as seen in the first column of every set in \refFig{DetailAlignment}.
When $N=28$ subjects were asked in a 2AFC to indicate ``which image shows a more correct alignment of the shape to the image'' they referred to ours in 59\,\% of the cases (significant $p<.0001$, binomial test), indicating that our approach is preferred and that computational metrics and human assessment are correlated.

\mysubsubsection{Qualitative}{QualitativeResults}

\mycfigure{DetailAlignment}{
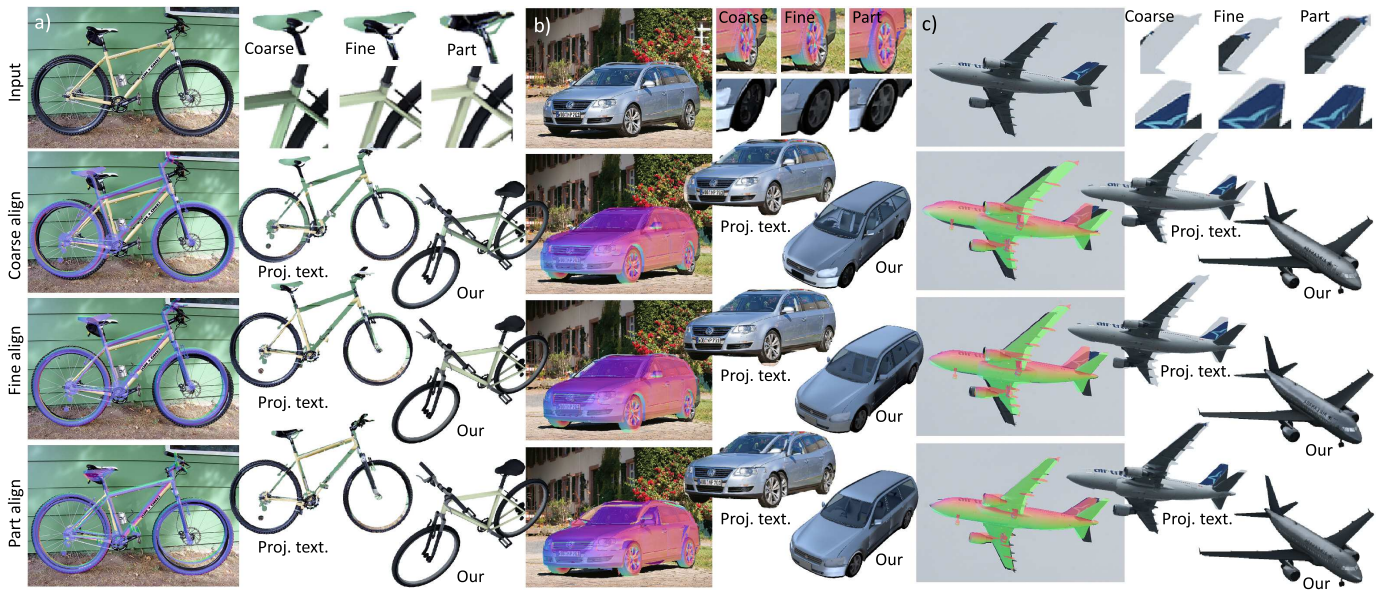
Our three stages of 2D-to-3D coarse-to-part alignment \emph{(horizontal, from the 2nd to the 4th row)} for different inputs \emph{(vertical)}. For every input, the 1st row shows the input image \emph{(1st col.)} and some insets taken from coarse \emph{(2nd col.)}, fine \emph{(3rd col.)} and part alignment \emph{(4th col.)} respectively. From the second row on: the 1st column shows the blending of the aligned mesh on top of the original image, the 2nd column shows projective texturing result and the 3rd column shows our directional-dependent appearance rendered from a different view.
}
Results of the alignment procedure are detailed in \refFig{DetailAlignment} and additional final renderings are shown  in \refFig{NovelViewResults}.
All results are obtained with minimal user input only from 2D images that we have obtained from the public Imagenet~\cite{Deng2009} database, for the classes indexed in Pascal3D \cite{xiang2014} dataset. We additionally provide results for the IKEA dataset~\cite{Lim13} using the images and their 3D models.
Our approach produces the alignment shown in less than 5\,s, whereas other work~\cite{Aubry2014} required many minutes on large clusters. For synthesizing the new views we use the view-dependent approach of~\refSec{ViewSynthesis2}.

\paragraph*{Alignment}
We detail the qualitative improvements achieved by each alignment step in \refFig{DetailAlignment}. For three input images of different classes (bike, station wagon, airliner), we show for each alignment step (coarse, fine, part) the overlaid 3D model in terms of surface normals, as well as novel view-renderings using projective texturing (left) as well as our Reflectance Map model (right). 

The first row shows in addition to the input image a couple of close-ups highlighting the obtained improvements in alignment and/or rendering quality. We would like to draw the reader's attention to a couple of improvements: For the bike example, we observe that starting from an overall greenish appearance of the bike, the fine alignment manages to recover an accurate representation, \eg in terms of a yellow frame and black saddle. The initial artifacts are due to bad alignment which corrupt the estimation of material appearances by green background pixels. The result shows the capability of our method to handle thin structure for which alignment is particularly difficult. 

For the car example, please note how the gradual alignment gets rid of background artifacts around the contour as well as dealing with transparent materials such as the glass windows. Finally, the airplane example, highlights the deformation component included in the fine alignment. While initially, the wings are not well covered due to object class variations in wing size and angle, the uniform scaling already achieves a closer match. In addition, we observe across all three examples, that the part-based alignment greatly improves on the fine alignment by accounting for non-linear variations. Note how the part-based stage handles variation in frame geometry and saddle height for the bike, variations in the wheel base for the car, position, size and angle of wings for airplanes.

\mycfigure{NovelViewResults}{
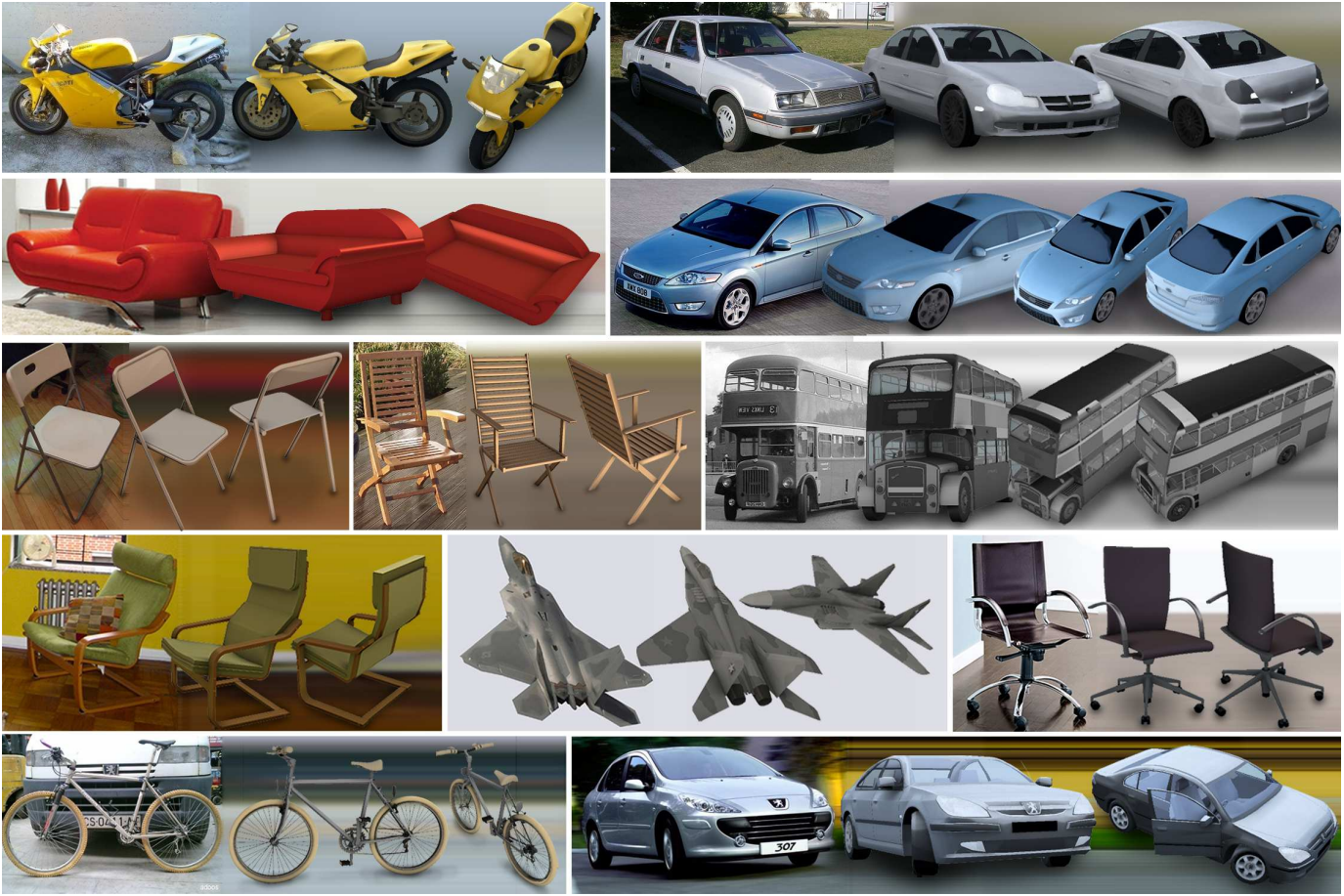
Novel views rendered by our algorithm in real-time after 5\,s of 2D-to-3D alignment for different 2D query images without manual intervention: input query image \emph{(1st column)} and results  \emph{(2nd and 3rd column)}.
Please see the project website for the respective videos. Images from Pascal3D\cite{xiang2014} and IKEA~\cite{Lim13}.
}

\paragraph*{Rendering}
\refFig{NovelViewResults} shows further results of the final rendering quality that highlight the truthful reproduction of complex appearances including specularities and transparency. In detail, we would like to point the reader's attention to the rendition of complex materials such as paint on cars and motorbikes that capture specular effects. Details that were not observed in the original view are rendered in the novel viewpoints matching the overall style of the object in a consistent manner. The rendering is consistent and seamless across all view points with no discernible change between the observed part and disocclusion areas from the input image.

Note that simple projective texturing cannot hallucinate the appearance in the disoccluded parts, especially for large viewpoint changes. Even if we use image inpainting (OpenCV) this method cannot capture the high-level structure of the object (\refFig{ProjectiveTexturing}).
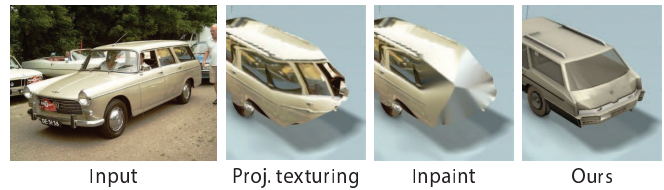
\myfigure{ProjectiveTexturing}{Effect of the projective texturing for a novel view. The area that is not visible in the initial view can not be synthesized properly because it contains high level structures that low level methods such as projective texturing or impainting can not capture. }

Our project website show novel-views for 390 random images from this database for 10 different classes from Imagenet and 3 furniture classes from the IKEA dataset\cite{Lim13}.
Please see the additional videos, for real-time screen-captures from a large selection of query images.

\mysection{Discussion}{Discussion}
When performing 2D-to-3D alignment without the fine and part-based steps to follow, either a very large number of detectors was required, or the alignment remained coarse.
Our alignment is both faster and higher quality than previous alignment methods for several reasons:
First, we need to evaluate a much lower number of templates, as the fine and part-based alignment can refine the solution.
Second, the evaluation of a single template is more efficient than evaluating multiple discriminative patches as it has been done in prior work \cite{Aubry2013}.
Yet, if the query object is too different from any object in the database, our approach produces a geometrically wrong object.
Still our appearance transfer is robust to mismatches between query and database object which cannot be avoided in practice.

Currently, we lift images with a single, unoccluded object to 3D.
Combining multiple objects is non-trivial and requires more than just detecting a lot of objects.
Combination with 3D inference that is not based on object-templates \cite{Hoiem2005,Saxena2009} would complement our approach.

The user intervention we require is minimal, but still it is required to indicate the object that should become 3D by drawing a 2D bounding box or scribbles.
We envision a future integrated system, where a background process finds objects in user content such as documents and webpages. Afterwards, interaction opportunities are advertised to the user where instant lifting to 3D of objects is possible.
The main prerequisite of our work is the requirement to indicate the location and class of the shape in the image using minimal interaction.
In future work, we want to combine our approach with fast and accurate object class recognition and localization (\eg, \cite{girshick2015fast}) as well as segmentation (\eg, \cite{long2014fully,he16arxiv}) and iterative pose refinement \cite{chiu15iccv}.
When a query image shows an object not in the database, a similar object is proposed, which may or might not be appropriate, depending on the application.
In future work, a combination with automatic or manual shape adjustments \cite{Kholgade2014} and more flexible shape representations \cite{sharma16arxiv} will further improve our coverage.

Our first method to generate novel views was presented in~\refSec{NovelViewSynthesis} and its extension in~\refSec{ViewSynthesis2}. The main difference between the two is where the appearance reconstruction is taking place. In the initial approach, every pixel in the novel image is synthesized directly as a weighted combination of the pixels in the original view, based on
the information from the aligned 3D model. In the second approach, the appearance reconstruction is performed through Reflectance Maps. From the  alignment, we reconstruct the Reflectance Maps by considering viewer-centered normals. Therefore, instead of reconstructing directly the appearance of pixels in the novel view, we reconstruct the appearance of normals in the initial view and we use this mapping to generate the novel views.

Both of the methods have advantages and disadvantages with respect to each other. The first approach allows the 3D position to participate in the reconstruction, enabling the effect of projective texturing in the visible area by giving a large value to the corresponding weight in~\refEq{ReconstructionAppear}. In this way we preserve the high frequencies for the visible parts, but in the disoccluded areas the reconstruction is blurred. This co-appearance of the two modalities in the same image is perceptually unpleasant.

Another important aspect is the task that we focus. In a user-manipulation scenario, we need the appearance to be view-dependent. This means that when we change viewpoint, the specularities and artifacts that appear in the object should change position. The second method considers view-dependent normals for the 3D model, thus after estimating the Reflectance Maps for each material, the new views will include this effect. Moreover, the complexity for the novel views is much lower, since the Reflectance Maps are essentially a look-up table of appearance for the surface normals of the 3D model. Otherwise, the first approach loops over all the pixels in the original view for every pixel in the new view.

Our system is not yet able to factor out reflectance, occlusion and lighting.
Furthermore, we do not support stochastic or structured reflectance changes such as textures or painted logos in the 2D query image.
Conversely, 3D shapes from on-line model repositories can contain details such as logos that are not part of the query image.
Telling apart such ``decoration'' from the essential 3D shape or 2D image deformation is an exciting direction of future research.

\mysection{Conclusion}{Conclusion}
We have proposed a system to hallucinate high-quality novel views from a single image containing an object of a known class at interactive rates.
Starting from the idea that output images can be written as a linear combination of input pixels where weights depend on the guidance of a manually aligned 3D model, we have devised an approach to generate novel views. Next, we have shown how to align the 3D model automatically, resulting in a second approach of novel-view synthesis that can produce results with minimal user intervention.
Applications include amplification of training data and almost every problem that requires novel views: stereography, synthetic motion blur and depth-of-field.
We have demonstrated that the appearance quality is good enough to improve performance in a classic detection and classification task.
Finally, the ability to mark an object in a 2D image and to lift it to 3D within just a few seconds \ie being able to inspect it from novel views, is an application in its own respect.
For future work, we would like to extend the synthesis and alignment to/from videos as well as to experiment with more or unknown object classes.


%

\ifCLASSOPTIONcaptionsoff
  \newpage
\fi



%
\bibliographystyle{IEEEtran}
\bibliography{IEEEabrv,arxivNovelView}

%

%
%


%




\end{document}